\documentclass{article}

\PassOptionsToPackage{numbers, compress}{natbib}
\usepackage[final]{neurips_2021}

\usepackage{lipsum}

\newcommand\blfootnote[1]{%
  \begingroup
  \renewcommand\thefootnote{}\footnote{#1}%
  \addtocounter{footnote}{-1}%
  \endgroup
}

\usepackage{microtype}
\usepackage{graphicx}
\usepackage{subfigure}
\usepackage{todonotes}
\usepackage{amsmath}
\usepackage{longtable}
\usepackage{xspace}
\usepackage[ulem={normalem,normalbf}]{changes}
\usepackage{soul}
\usepackage{svg}
\usepackage{booktabs} % for professional tables
\usepackage{amsthm}
\usepackage{amssymb}
\usepackage{wrapfig}
\usepackage[font={small,it}]{caption}
\usepackage{floatrow}
\newfloatcommand{capbtabbox}{table}[][\FBwidth]
\usepackage{hyperref}

\definecolor{reddish}{rgb}{0.5, 0.0, 0.3}
\definechangesauthor[name={Sebastian Jaszczur},color=reddish]{SJ}
\definechangesauthor[name={Afroz Mohiuddin},color=red]{AM}
\definechangesauthor[name={Henryk Michalewski},color=pink]{HM}
\definechangesauthor[name={Aakanksha Chowdhery},color=blue]{AC}

\newcommand{\pparagraph}[1]{\paragraph{#1}}
\newcommand{\ourname}[0]{Scaling \xspace}
\newcommand{\ourmodel}[0]{Terraformer}

\title{Sparse is Enough in Scaling Transformers}

\author{%$$
  Sebastian Jaszczur$^{*}$\\
  {University of Warsaw}\\
  \And
  Aakanksha Chowdhery\\
  {Google Research}\\
  \And
  Afroz Mohiuddin\\
  {Google Research}\\
  \And
\L{}ukasz Kaiser$^{*}$\\
{OpenAI}\\
\And
Wojciech Gajewski\\
{Google Research}\\
\And
Henryk Michalewski\\
{Google Research}\\
\And
Jonni Kanerva\\
{Google Research}\\
}

\newcommand{\dm}{d_\textrm{model}}
\newcommand{\dff}{d_\textrm{ff}}

\begin{document}

\maketitle

\begin{abstract}

\blfootnote{${}^{*}\textrm{Work done while at Google Research.}$}\hskip -0.05em
Large Transformer models yield impressive results on many tasks, but are expensive to train, or even fine-tune, and so slow at decoding that their use and study becomes out of reach. We address this problem by leveraging sparsity. We study sparse variants for all layers in the Transformer and propose \emph{\ourname Transformers}, a family of next generation Transformer models that use sparse layers to scale efficiently and perform unbatched decoding much faster than the standard Transformer as we scale up the model size. Surprisingly, the sparse layers are enough to obtain the same perplexity as the standard Transformer with the same number of parameters. We also integrate with prior sparsity approaches to attention and enable fast inference  on long sequences even with limited memory. This results in performance competitive to the state-of-the-art on long text summarization.

\end{abstract}

\section{Introduction} \label{intro}

The field of natural language processing has seen dramatic improvements in recent years
due to large neural networks based on the Transformer architecture. The original Transformer
\citep{vaswani2017attention} significantly advanced state-of-the-art in machine translation.
BERT \citep{devlin2018bert} surpassed all previous methods on question answering, language inference
and other NLP tasks and was followed by a line of models like T5 \citep{raffel2020exploring} that further
improved these results. The GPT line of models \citep{gpt2,brown2020language} elevated language
generation to the point that GPT-2 was invited to write short passages for the Economist and GPT-3
created whole articles almost indistinguishable from human-written ones.

The benefits of this progress are undercut by the huge costs such models incur. \citet{energyuse} 
estimate that training a single base BERT model costs \$$4$k-\$$12$k and emits as much CO$_2$
as one passenger's share of a $4$-hour flight  and later \citet{patterson2021carbon} estimate that training GPT-3 has three times as much tCO$_2$e (metric tons of CO$_2$ equivalent) emissions as a SF-NY round trip flight. 
Data and serving costs are also forbidding: a single
training run of BERT, for example, processes 128B tokens, and Google Translate
reportedly\footnote{\url{https://cutt.ly/skkFJ7a}} serves over 143B words per day.

With the growing popularity and size of these models, it is increasingly valuable to make them
scale efficiently. In this work we propose \emph{\ourname Transformers} with a separate {\bf\it sparse mechanism for the query, key, value and output layers} (QKV layers for short)
and combine it with {\bf\it sparse feedforward blocks} to get a fully sparse Transformer architecture.

To quantify the computational complexity of inference in Transformer models, recall the architecture
of a Transformer decoder block. It consists of three parts: a masked self-attention layer, an encoder-decoder
attention layer and a feedforward block. The sizes of these layers are parameterized by
$\dm$ and $\dff$. The base BERT model sets $\dm = 768$, the large BERT has $\dm = 1024$, the largest
GPT-2 has $\dm = 1600$ and GPT-3 reaches $\dm = 12288$. For both BERT and GPT models the authors 
use $\dff = 4 \, \dm$.
While decoding a token, the self-attention layer needs to activate four matrices of
size $\dm \times \dm$: one each for the queries, keys and values input to the attention and one for
merging the output. In the encoder-decoder attention, the keys and values may already be cached,
so only two matrices of size $\dm \times \dm$ are activated. The feedforward block consists of two matrices
of size $\dm \times \dff$, omitting small additional contribution of biases. The total adds up to:
%\[ 
$4 \, \dm^{\,2} + 2 \, \dm^{\,2} + 2 \, \dm \, \dff .$
%$\]
This sum describes both the number of trainable weights of a single block and approximates well
the number of floating-point operations needed for decoding a single token, except for the attention
operations (discussed later). The complexity is quadratic in $\dm$; for example, as $\dm$ increases
$16$-fold from base BERT to GPT-3, the complexity of a single block grows $256$-fold.

\begin{figure}
\begin{floatrow}
\capbtabbox{%
\resizebox{0.475\textwidth}{!}{
\begin{tabular}{r  c  c c} 
\hline
  & Params & Dec. time & Dec. time \\
  & & & per block\\
 \hline
 baseline Transf. & 800M& 0.160s & 5.9ms \\ 
 + Sparse FF  & - & 0.093s & 3.1ms \\
 + Sparse QKV  & - & 0.152s & 6.2ms \\
 %\hline
 + Sparse FF+QKV & - & 0.061s & 1.9ms \\
 \hline
 Speedup  & & 2.62x & 3.05x \\
 \hline

 baseline Transf. & 17B & 3.690s & 0.581s \\
 %\hline
 +Sparse FF & - & 1.595s  & 0.259s \\
 +Sparse QKV & - & 3.154s  & 0.554s \\
 %\hline
 +Sparse FF+QKV  & -& 0.183s & 0.014s \\  
 \hline
 Speedup & & 20.0x & 42.5x \\
  \hline
\end{tabular}}
}{%
\caption{Decoding speed (in seconds) of a single token. For Transformer model (equivalent to T5 large with approximately 800M parameters), \ourname Transformers with proposed sparsity mechanisms (FF+QKV) achieve up to 2x speedup in decoding compared to baseline dense model and 20x speedup for 17B param model. 
}
\label{table:intro_sparse_overall}
}
\ffigbox{%
    \centering
    \includegraphics[width=0.5\textwidth,keepaspectratio]{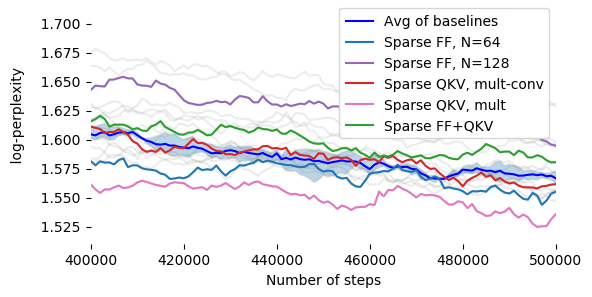}
}{%
    \caption{Log-perplexity of \ourname Transformers (equivalent to T5 large with approximately 800M parameters) on C4 dataset with proposed sparsity mechanisms (FF, QKV, FF+QKV) is similar to baseline dense model. Other models used in this paper are shown in grey lines; raw data is available in the appendix.}%
    \label{fig:intro_sparse_overall}}
% }
\end{floatrow}
\end{figure}

In comparison  \emph{\ourname Transformers} use
only $2 \dm \sqrt{\dm} = 2\dm^{\,1.5}$ parameters in QKV layers and yield results as good as the baseline (fully dense) Transformer with the same number of parameters and complexity:
%\[
$8 \, \dm^{\,1.5} + 4 \, \dm^{\,1.5} + \, 4 \, \dm^{\,1.5} .$
%\]
We were surprised that the fully sparse \emph{\ourname Transformers}
% model is
are indeed enough to match the results of the baseline Transformer on the large C4 dataset \cite{raffel2020exploring} (Figure~\ref{fig:intro_sparse_overall}).
The improvement in complexity holds not just asymptotically but yields over 2.6x speedup in wall-clock  hed
decoding time already for a model with 800M parameters and 20x improvement for a model with 17B parameters,
as shown in Table~\ref{table:intro_sparse_overall}.

\footnotetext[2]{The 800M model has 24 layers of Encoder \& Decoder, $\dm=1024$, $16$ attn heads, $\textrm{attention-sparsity}=16$, $\textrm{ff-sparsity}=64$. We scale up this model to approximately 17B parameters with $\dm=9216$ and get up to 20x speedup in decoding compared to baseline dense model. This 17B param model has six layers of Encoder \& Decoder, $96$ attn heads, $\textrm{attention-sparsity}=64$, $\textrm{ff-sparsity}=256$.}

To verify that Scaling Transformers can be used with other Transformer improvements on real tasks, 
we create \emph{Terraformer} -- a Transformer model that uses reversible layers for memory efficiency
and sparse attention to handle long sequences. We pre-train Terraformer on the C4 dataset and fine-tune it on the challenging task of summarizing arxiv articles. Terraformer yields results competitive to the state-of-the-art BigBird-Pegasus without using the Pegasus loss in pre-training (Table~\ref{table:arxiv}).

\section{Related Work}

As discussed in the previous section, large Transformer models brings significant improvements in performance, as seen in models such as GPT-3~\cite{brown2020language, kaplan2020scaling} or T5~\cite{xue2020mt5,raffel2020exploring}. Training and inference incur a high computational cost at the scale of hundreds of billions of parameters. Numerous techniques improve the efficiency of Transformer models, and \citet{gupta2020compression} divide them into several classes, including pruning, knowledge distillation, quantization, parameter sharing, efficient attention, and efficient feedforward.

\textbf{Model compression.} Model pruning~\cite{li2020train, brix2020successfully} makes matrices smaller by removing unneeded weights after or during training, 
however, the gains in computational complexity on sparse matrices
often do not result in inference speedups on actual hardware~\cite{gale2019state}. Structured pruning based approaches \cite{zhou2021learning, li2020efficient, wang2020hat} account for this challenge by leveraging sparsity in hardware in CPU and GPU architectures~\cite{NvidiaAmpereArch}. 
Our paper is different from pruning approaches in that it relies on dynamic sparsity wherein the feedforward layer loads only a subset of weights in the layer for each token. Our approach is complementary to model quantization studies~\cite{shen2020q, sun2019hybrid, prato2019fully} that use fewer bits for the weights.

\textbf{Model distillation.} Several natural language
models used for mobile inference~\cite{iandola2020squeezebert,sun2020mobilebert} rely on distillation \cite{sanh2019distilbert} to speed up inference from the pretrained large models. For example, \cite{kim2020fastformers} pretrains a large model and uses knowledge distillation along with pruning to get more than 10x faster inference. Instead of distilling a large model, our approach speeds up inference by reducing the number of weights loaded in memory from the model.

\textbf{Sparse attention.} Sparse attention-based approaches have made the attention layer more efficient, especially for long sequences, by incorporating additional combinatorial mechanisms, as in \cite{tay2020sparse},
or selecting a subset of tokens this layer attends to \cite{roy2020efficient,choromanski2020rethinking, kitaev2020reformer, sukhbaatar2019adaptive, kb2018discrete, child2019generating} or other approaches \cite{lwc18}. Our work is complementary to these
approaches for sparse attention and reuses the advances on SOTA therein. 
Inference speedups in the attention layers also use bottleneck layers~\cite{sun2020mobilebert} or grouped convolutions~\cite{iandola2020squeezebert}. Our work extends beyond the idea of grouped convolutions approach because each attention head is limited to
using only a fixed part of the embedding while our work is able to permute the embeddings to improve model quality; see Section~\ref{sec:sparse-qkv} for details.

\textbf{Tensor Decomposition.} The approaches discussed above significantly improve Transformer speed and handling of long sequences, 
\hbox{however} none of them addresses the fundamental scaling issue: even if we distill into a smaller model,
quantize it and prune a percentage of the weights, the complexity still grows quadratically
with $\dm$. The final approach, which does attack this scaling issue, is called \emph{tensor decompositions} in \cite{gupta2020compression}.  Unluckily, as the authors there note, the approach is most effective in
dealing with large input and output embedding matrices and tends to produce lower performance than
unstructured models if used inside the decoder block. 

\textbf{Sparse feedforward.} 
Mixture of experts approaches have been shown to achieve computational efficiency in training \cite{shazeer2017outrageously, lepikhin2020gshard, shazeer2018mesh}, scaling up to a trillion parameters
\cite{switchtransformer}.
The key idea is to partition the $\dff$-sized dimension into parts (called experts) and retrieve only one part per token, which reduces the complexity of the feedforward block from $2\dm\dff$ to $2\dm\dff / n_\textrm{experts}$. 
These speedups are mostly measured in training speed, and the method focuses on feedforward blocks.
In contrast to prior methods, we train a full weight matrix and then only activate specific parts of it for each input token during decoding; 
see Section~\ref{sec:sparse-ff}.

\section{Sparse is Enough}

We study how to sparsify every part of the Transformer model---otherwise the non-sparse parts dominate decoding time and become a bottleneck. This means we need sparse equivalents for the feedforward blocks, for the dense Q, K, V and output layers in attention, and for the final dense layer before the softmax and loss.

\subsection{Sparse Feedforward Layer}
\label{sec:sparse-ff}
In a baseline Transformer, decoding speed is dominated  by the execution cost of
the feedforward block. Recall that this block consists of two fully-connected (dense) layers
with a ReLU nonlinearity in between. The dimensionality of activation vectors
between these 2 layers is usually denoted by $d_{\text{ff}}$ and is often 4 or 8 times larger than
the dimensionality of the activations in other places ($d_{\text{model}}$).

\begin{figure*}
    \centering
    \subfigure[]{%\includesvg[width=0.52\textwidth]{images/SparseFFMain3.svg}}
    \includegraphics[width=0.52\textwidth]{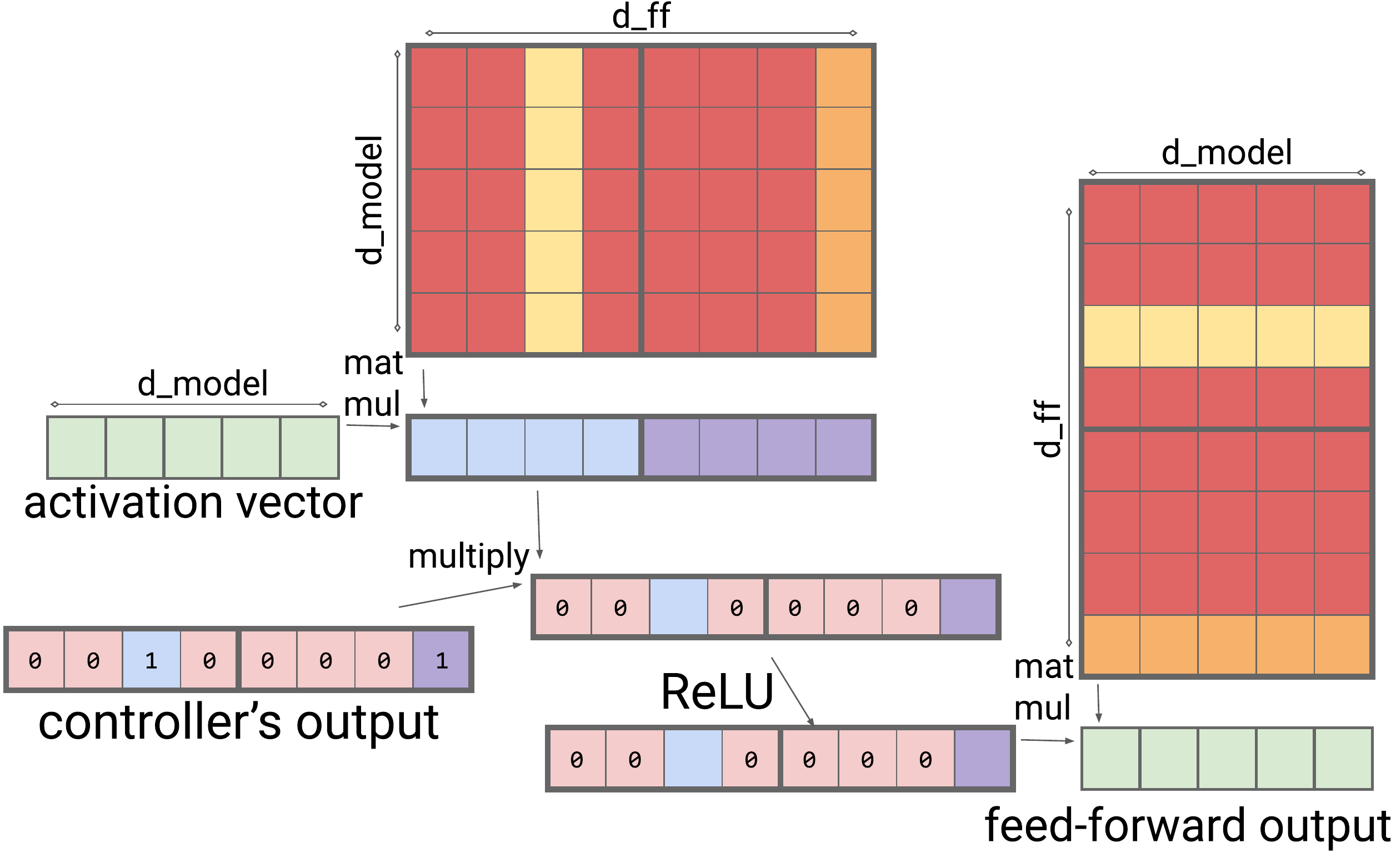}} 
      \hspace{1mm}
    \subfigure[]{%\includesvg[width=0.44\textwidth]{images/SparseFFController2.svg}}
    \includegraphics[width=0.44\textwidth]{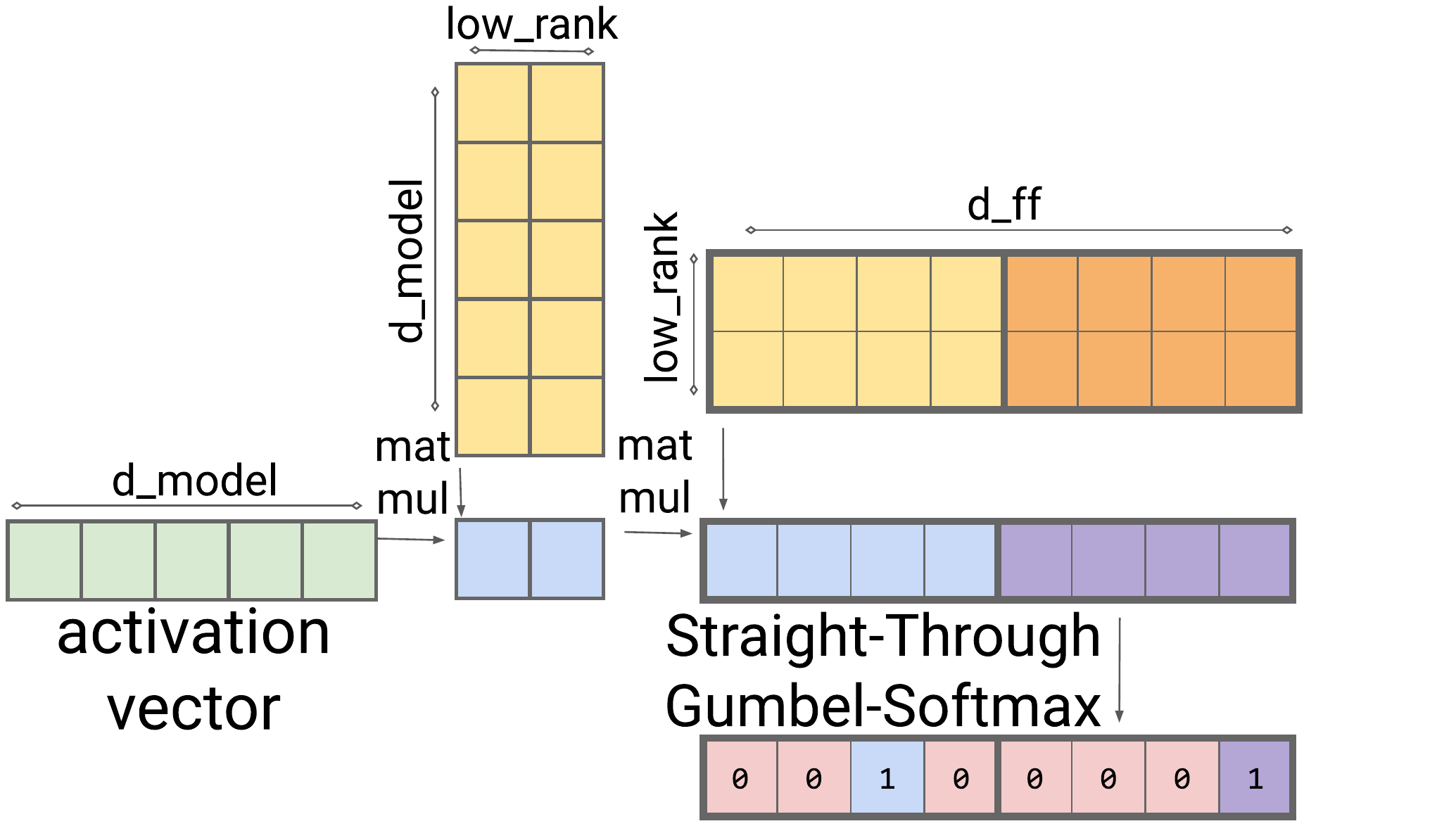}} 
    \caption{(a) Sparse Feedforward Layer only activates 1 in N rows/columns of each block to reduce the decoding time. Here only two rows/colums in blocks of size 4 are loaded while the weights in dark red are not loaded from memory during inference. (b) Sparse Feedforward Controller with the output of 2 blocks of size 4 (1 in 4 sparsity).}
    \label{fig:SFFController2}
\end{figure*}
        
We make use of the structure of the feedforward block to sparsify it. One main observation is that
the ReLU in the middle creates a lot of zeros\footnote{GeLU is another non-linearity often used in the Transformer feedforward block. Table 1 in \citep{tmods} shows the same final loss for ReLU and GeLU on the C4 dataset, though, so in this work for simplicity, we focus on ReLU.}. We impose a fixed structure on this middle activation vector: only one float in every block of $N$ will be allowed to be non-zero. Prior techniques prune weights or blocks from weight matrices and can be referred to as static sparsity. Our proposed technique will train a full weight matrix but only activate specific parts of it for each input token during decoding. We call this dynamic sparsity, because the model dynamically selects only a fraction of its parameters, and the selection is independent for each token. 

We train a controller to determine which activation in each block can be non-zero; the rest will be set to zero. This can be represented as
\begin{gather*}
    Y_\textrm{sparse} = \max(0, x W_1 + b_1) \odot \textrm{Controller}(x) \\[.5ex]
    \textrm{SparseFFN}(x) = Y_\textrm{sparse} W_2 + b_2
\end{gather*}
where $\odot$ is element-wise multiplication. Note that each activation in $Y_\textrm{sparse}$ corresponds to a single column in $W_1$ and a single row in $W_2$. Therefore, if we compute $\textrm{Controller}(x)$ output first, we don't have to use any columns in $W_1$ or any rows in $W_2$ that correspond to an activation set to zero by the controller. This allows for much faster decoding, as we have to process only 1 in $N$ columns in $W_1$ and rows in $W_2$ (see Figure \ref{fig:SFFController2}(a)).

To design the controller to be computationally inexpensive, we project the input using a low-rank bottleneck dense layer.
Figure~\ref{fig:SFFController2}(b) illustrates the controller which produces the output as follows
\[
    \mathrm{Controller}(x) = \mathrm{arg\,max}(\mathrm{Reshape}(x C_1 C_2, (-1, N)))
\]
where $C_1 \in \mathbb{R}^{d_\textrm{model} \times d_\textrm{lowrank}}$ and $C_2 \in \mathbb{R}^{d_\textrm{lowrank} \times d_\textrm{ff}}$, with $d_\textrm{lowrank}$ usually set to ($d_\textrm{model}/N$).

During inference the controller uses a discrete argmax function, but during training the model uses a softmax to calculate and sample from a distribution. The model learns to select which row/column will be non-zero using the Gumbel-Softmax trick for discretization. To determine the active row/column in each block, we reparameterize sampling from a Bernoulli distribution by using the Gumbel-Softmax trick ~\cite{maziarz2019gumbel}. 
Instead of using the logits in each block to directly sample a binary value, we add independent noise from the
Gumbel distribution to each of the logits, and then select the binary value with the highest logit (i.e.,
argmax) as the sample $z$. 
The argmax operation is not differentiable, but it can be approximated by a softmax with annealing
temperature. Therefore, on the forward pass, we use the argmax to obtain a binary one-hot vector for each block,
while on the backward pass, we approximate it with softmax. This
approach is known as the Straight-Through Gumbel-Softmax estimator \cite{jang2016categorical}.

\begin{figure}
\begin{floatrow}
\capbtabbox{%
% \begin{table}
% \begin{center}
% \centering
\begin{tabular}{r c} 
\hline
    & Dec. time\\
 \hline
 baseline  & 0.160s \\
 %\hline
 %\hline
 Sparse FF 64  & 0.093s  \\
 %\hline
 Sparse FF 128 & 0.089s  \\
  \hline
\end{tabular}
% \end{center}
}{%
\caption{Decoding time of a singe token decreases with increasing level of sparsity in the FF layer. }
\label{table:sparse_ff_ablation}
}
\ffigbox{%
    \centering
    \includegraphics[width=0.5\textwidth,keepaspectratio]{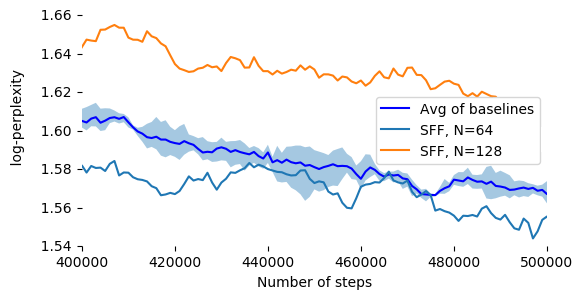}
    % \vspace{-0.1in}
}{%
    \caption{Log-perplexity of \ourname Transformers with Sparse Feedforward layer is very similar to dense baseline for sparsity level $N=64$ but degrades slightly for N=128.}%
    \label{fig:sparse_ff_ablation}
    % \vspace{-0.2in}
}
    
\end{floatrow}
\end{figure}

\pparagraph{Ablations.} We investigate the impact of sparse FF on the model equivalent to T5-large with varying levels of sparsity,  with $d_\textrm{model} = 1024$, $d_\textrm{ff} = 4096$, and 16 attention heads. When we set the sparsity level to $N$ (for e.g. $N=64$) then every block of size $N$ has one non-zero value activated for inference. During training, the controller uses the bottleneck layer with $d_\textrm{lowrank} = 64$ and temperature of  Gumbel softmax estimator set to 0.1. To improve training stability, the controller in the forward pass will use the output of argmax that is a binary one-hot vector for each block with a probability of 30\% and otherwise it uses the output of softmax.
Table \ref{table:sparse_ff_ablation} and Figure~\ref{fig:sparse_ff_ablation} show the perplexity and the decoding time of this model with varying levels of sparsity in feedforward layer. As the level of sparsity increases from 0 to 128, we observe a significant decrease in the decoding time, while the neg-log-perplexity of the model with $N=64$ sparsity is comparable to the baseline.

We also checked the performance of the feedforward block with Mixture-of-Experts \cite{shazeer2017outrageously} style sparsity. As expected, this technique achieved decoding time comparable to sparse FF -- 0.11s instead of 0.09s -- but with its lack of granularity it achieved log-perplexity of 1.64, worse than both our method and the dense baseline.

%     \centering
%\begin{minipage}{0.4\textwidth}
% \end{figure}
% \end{minipage}

\subsection{Sparse QKV Layer}
\label{sec:sparse-qkv}
The decoding speed for a model with sparse feedforward blocks is dominated next by
the query, key, value and output computation---the dense layers in attention, which we jointly call a
QKV layer. Each of these dense layers has $\dm^{\,2}$ parameters and computation cost.
Unfortunately, QKV layers don't have ReLUs, so the method used above to sparsify feedforward blocks is
not viable here. 

\begin{figure*}
    \centering
    \subfigure[]{%\includesvg[width=0.48\textwidth]{images/Multiplicative2.svg}
    \includegraphics[width=0.48\textwidth]{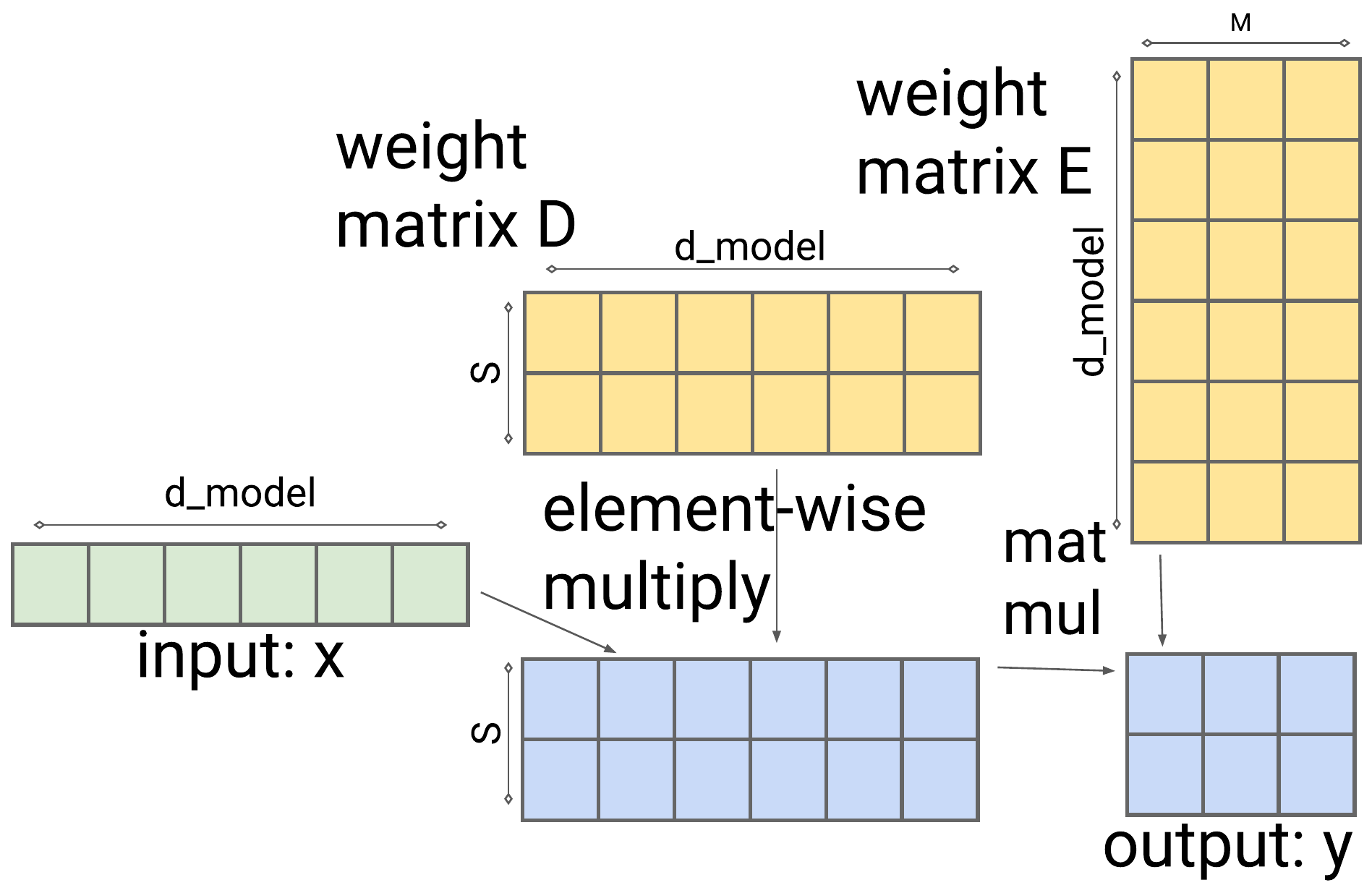}} 
      \hspace{1mm}
    \subfigure[]{%\includesvg[width=0.48\textwidth]{images/SparseQKV2.svg}
    \includegraphics[width=0.48\textwidth]{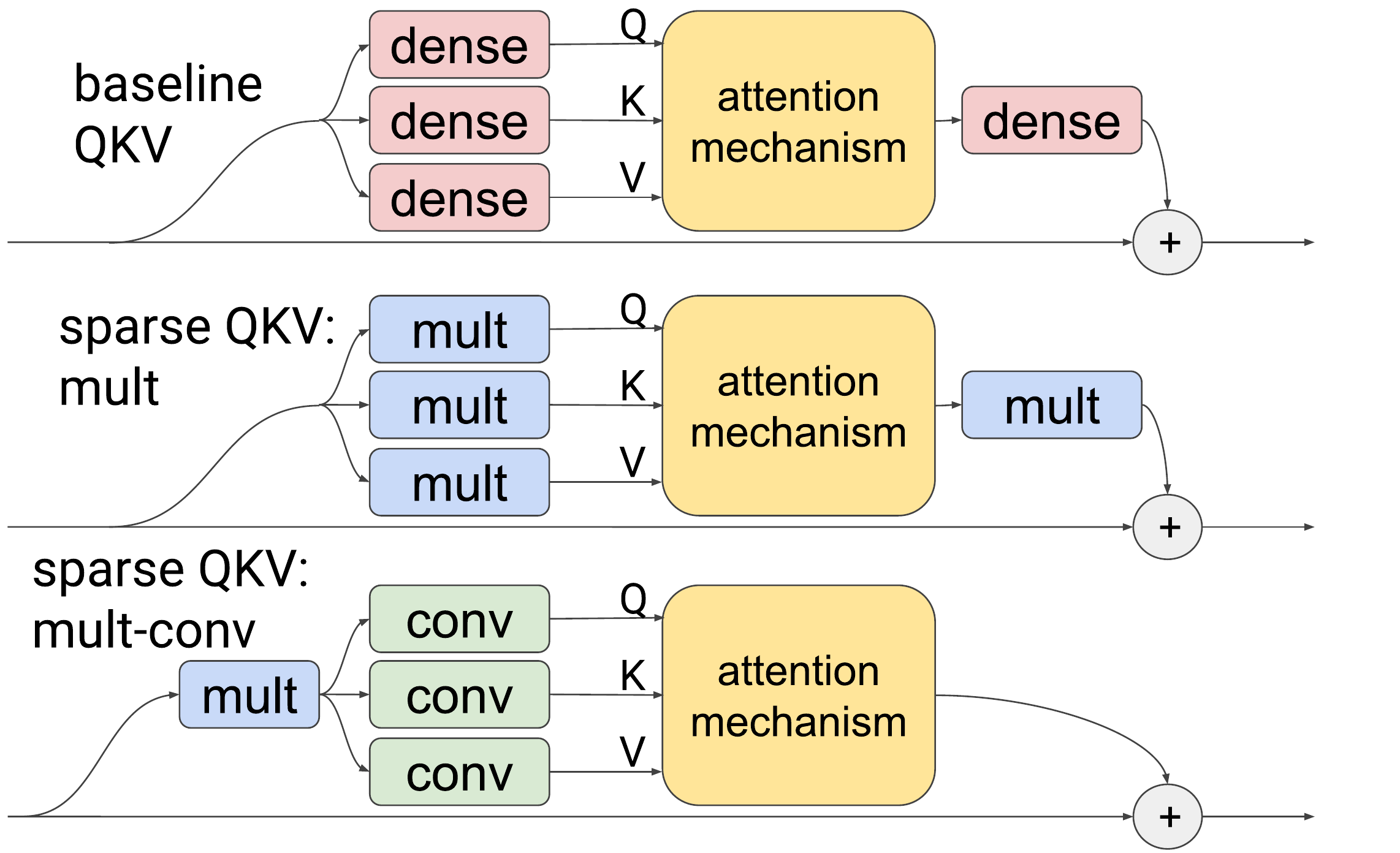}} 
    \caption{(a) Multiplicative layer can represent an arbitrary permutation, but has fewer parameters and reduced computation time compared to a dense layer. (b) Sparse QKV layer replaces Q, K, and V dense layers by composing multiplicative and convolutional layers and reducing the number of parameters and decoding time.}
    \label{fig:SparseQKV}
\end{figure*}

To make QKV layers sparse, we subdivide the dimensionality of the layer, $\dm$, into
$S$ modules of size $M = \dm/S$, similar to splitting an activation vector into multiple heads. These modules can be processed with a convolutional layer with fewer weights and faster computation. However, with na\"ive design each module (and corresponding attention head) could access only a small part of a given token embedding. To alleviate that,
we develop a multiplicative layer that can represent an arbitrary permutation and has fewer parameters and lower computation time than a dense layer. This multiplicative layer is inserted right before the convolutional layer, letting each head access any part of the embedding (see Figure~\ref{fig:SparseQKV}(a)). This solution yields well-performing models that also decode fast.

\paragraph{Multiplicative dense layer.}

Our new multiplicative dense layer can represent an arbitrary permutation and has $\dm^{\,2}/S + \dm S$ parameters, dependent on the sparsity hyperparameter $S$. It processes an input vector $\textrm{x} \in \mathbb{R}^{\dm}$ by splitting it into S ``modules'' of size $M = \dm/S$. 
It produces output $\textrm{y} \in \mathbb{R}^{S \times M}$ as follows
\[
    \textrm{y}_{s, m} = \sum_{i} \textrm{x}_i D_{i,s} E_{i,m}
\]
where the two weight matrices are $ D \in \mathbb{R}^{\dm \times S}$, and $ E \in \mathbb{R}^{\dm \times M}$ (see Figure~\ref{fig:SparseQKV}(b)). This layer executes significantly faster during inference because of the decreased number of parameters which need to be loaded from memory. Unless stated otherwise, we use $S=16$.
\newtheorem{theorem}{Theorem}

The multiplicative layer is designed primarily to represent any permutation, so that each attention head can access information from any part of the embedding. We first verify that the multiplicative layer can indeed represent an arbitrary permutation (the proof is presented in the Appendix).
\begin{theorem}
For any bijective function $f: \{1 \cdots \dm \}\Rightarrow \{1 \cdots S\} \times \{1 \cdots M\}$ there exists a pair of weights of multiplicative layer D, E such that $x_i = y_{s,m}$ for $\{s,m\} = f(i)$.
\end{theorem}

\pparagraph{Convolutional layer.}
The output of the multiplicative layer is a tensor of type/shape $\in \mathbb{R}^{\textrm{batch} \times \textrm{length} \times S \times M}$. We process this tensor with a two-dimensional convolutional layer, treating the length dimension and number of modules $S$ like height and width of an image. This layer uses $M$ filters and a kernel size of $F \times F$ so that each filter looks at $F$ modules (`S' axis) of the last $F$ tokens (`length' axis). Replacing the standard dense layer with such a convolution reduces the parameter count and computation time of the QKV layer. At the same time, by convolving over the `length' axis, the model can incorporate more context into this computation \cite{li2019enhancing}.

The output of this layer has the same shape as the input. The optimal value of $S$ is less than $\sqrt{\dm}$. Empirically we set $F$ to $3$, $S$ equal to the number of heads in the attention mechanism and $M$ to be the dimensionality of a single attention head. In this case, we can feed the output of the convolution directly to the attention mechanism without reshaping the output. This convolutional layer has fewer parameters ($9M^2 + M = F^2(\dm/S)^2 + (\dm/S)$), and lower computational complexity ($O(\dm^{\,2}/S)$). Unless stated otherwise, we use $S=16$ and $F=3$.

\pparagraph{Combining multiplicative and convolutional layers.} There are four dense layers to replace in the original attention mechanism: Q, K, V, and output. As shown in \hbox{Figure}~\ref{fig:SparseQKV}(b), we replace Q, K, and V dense layers by composing multiplicative and convolutional layers, but with a multiplicative layer shared across all three: $Q=\textrm{conv}_Q(\textrm{mult}(x))$, $K=\textrm{conv}_K(\textrm{mult}(x))$, $V=\textrm{conv}_V(\textrm{mult}(x))$. We remove the output dense layer. Note that the combined multiplicative-convolutional variant has the output dense layer removed, while the other variants have it replaced with their respective sparse layers. Including this output layer negatively impacts decoding time. We can set the parameter $S$ to around $\sqrt{d_{model}}$, getting the number of layer parameters to scale proportionally to $d_{model}^{\,1.5}$ compared to $d_{model}^{\,2}$ of standard QKV layer.

\pparagraph{Interpretation of QKV layer.} Note that when parameter $S$ in convolutional layer is equal to the number of heads in the attention mechanism, which is the case in our experiments, then each of the S modules corresponds to a single attention head. Therefore, the model uses the convolution to process each head using the same linear projection. Without the multiplicative layer this projection would operate on a predetermined part of the embedding layer for each head. However, by adding it the model can perform arbitrary permutation of dimensions, so each head can have access to arbitrary subset of embedding dimensions, not a predetermined subset of them. This fact helps with keeping the expressibility of resulting QKV layer despite the reduced number of parameters.

\begin{figure}
\centering
    \begin{minipage}{0.6\textwidth}
    \centering
    \includegraphics[width=1\columnwidth,keepaspectratio]{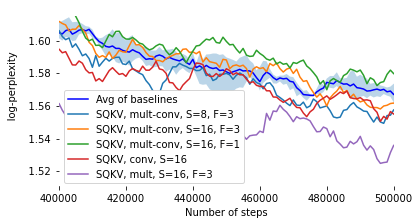}
    \vspace{-0.1in}
    \caption{Log-perplexity of \ourname Transformers with Sparse QKV with different sparsity levels (S) and kernel sizes (F) is very similar to dense baseline within variance while multi-layer even improves perplexity.}
    \label{fig:sparse_qkv_variants}
    % \vspace{-0.15in}
    \end{minipage}
\end{figure}

\pparagraph{Ablations.} We investigate the impact of sparse QKV layers on the model equivalent to T5-large in Figure~\ref{fig:sparse_qkv_variants}. We increase the value of $\dff$ from 4096 to 6144 to preserve the number of parameters (see the next subsection for details). The decoding time with sparse QKV layer variants is similar to the baseline because it is dominated by the dense feedforward layer (details in appendix).

\pparagraph{Combined feedforward and QKV sparsity.}
Sparse QKV layers lower the total number of model parameters. To keep the model size matched to the baseline, we increase $\dff$ to keep the number of parameters similar across all models we compare. For the T5-Large equivalent model, we increase $\dff$ from $4096$ to $6144$. With increased $\dff$, decoding time in the feedforward layer increases and thus, Sparse QKV layers alone do not speed up the model. However, when we combine Sparse QKV layers with sparse FF layers, we get a 3.05x speedup in decoding time of each decoding block with comparable perplexity (see Table~\ref{table:intro_sparse_overall} and Figure~\ref{fig:intro_sparse_overall}). While the baseline these is a vanilla Transformer, the decoding speed is almost the same for a Reformer model as well.

Table~\ref{table:glue_sparse_ff_qkv} shows the accuracy of fine-tuning the model for downstream tasks from the GLUE dataset. Note that the model with sparseFF+QKV achieves accuracy similar to the baseline.

\subsection{Sparse loss layer.}
A final dense layer maps the model embedding into vocabulary size to compute the loss.
We can sparsify this part of the model by replacing the dense layer with
a multiplicative layer similar to previous sections; this speeds up decoding time but may degrade perplexity. 
The results are presented in appendix.

\begin{table*}
\resizebox{1\textwidth}{!}{
\centering
\begin{tabular}{r c c c c c c}
\hline
& RTE & MRPC & SST-2 & QNLI & MNLI-m & QQP \\ %[0.5ex] 
 \hline
Baseline Transformer (dense) & $70.1\pm 1.1$ & $83.6\pm 0.72$ & $92.6\pm 0.85$ & $88.6\pm 0.5$ & $78.5\pm 0.41$ & $85.2\pm 0.6$\\ 

 Scaling Transformer (Sparse FF+QKV) & $68.4$ & $81.2$ & $91.6$ & $90.1$ & $82.9$ & $89.9$\\ 
 Terraformer (Sparse FF+QKV) & $66.1$ & $84.6$ & $92.3$ & $88.3$ & $79.1$ & $85.5$ \\
\hline
\end{tabular}}
\caption{Accuracy of Scaling Transformer model and Terraformer model with sparse QKV+FF is comparable to the baseline Transformer within variance. The results are obtained by fine-tuning on selected downstream tasks from the GLUE dataset (validation split).}
\label{table:glue_sparse_ff_qkv}
\end{table*}

\section{Sparsity for Long Sequences}
\label{sec:terraformer}

The above gains from sparsifying the dense layers are encouraging,
but we omitted one fundamental issue. When applied to longer sequences,
the gains would effectively be lost, as the decoding time will be dominated by
attention operations. Luckily, a number of methods have been proposed
to solve this problem for Transformers, see \cite{tay2020efficient} for a survey.
We focus on the LSH (Locality-Sensitive Hashing) attention from Reformer \cite{kitaev2020reformer} and
show how to integrate this sparse attention mechanism, as well as recurrent
blocks, into a \ourname Transformer, yielding a \emph{\ourmodel}.

\subsection{Architecture for Long Sequences}
\label{sec:terraformer-arch}

While integrating sparse attention layers into a Scaling Transformer, we notice that the architecture of the Transformer decoder block is suboptimal and can be redesigned to make a better use of these layers. In particular, separating decoder self-attention and encoder-decoder attention is not necessary any more from the perspective of efficiency. We therefore remove the encoder-decoder attention, but just concatenate the encoder representations before the decoder tokens. Doing this alone isn't enough though, since we took away one attention mechanism (encoder-decoder attention). We remedy this by having two attention mechanisms before the feedforward block. This simple architecture is as fast as the baseline Transformer while giving better results.

Putting this together, if $v_{enc}$ are the encoder activations and $v_{dec}$ are the decoder embeddings,
the input to the decoder block $x$ is their concatenation on the length axis,
$\textrm{LengthConcat}(v_{enc}, v_{dec})$. Each decoder block can be represented as: 
\begin{align*}
    y_1 &= \;x + \textrm{Dropout}(\textrm{Attention}(\textrm{LayerNorm}(x))) \\[.5ex]
    y_2 &= y_1 + \textrm{Dropout}(\textrm{Attention}(\textrm{LayerNorm}(y_1))) \\[.5ex]
    y   &= y_2 + \textrm{FFN}(y_2)
\end{align*}
where $y$ becomes the input to the next decoder layer.
See the appendix for a full diagram of the resulting architecture.

\subsection{Reversibility for Memory Efficiency}
\label{sec:terraformer-reverse}
To enable training Terraformer with large batches, and to fine-tune even large models on single machines,
we apply ideas from the Reformer \cite{kitaev2020reformer}, in particular, reversible layers for
the encoder and decoder blocks.

\begin{figure}
    \centering
    \includegraphics[width=.7\columnwidth]{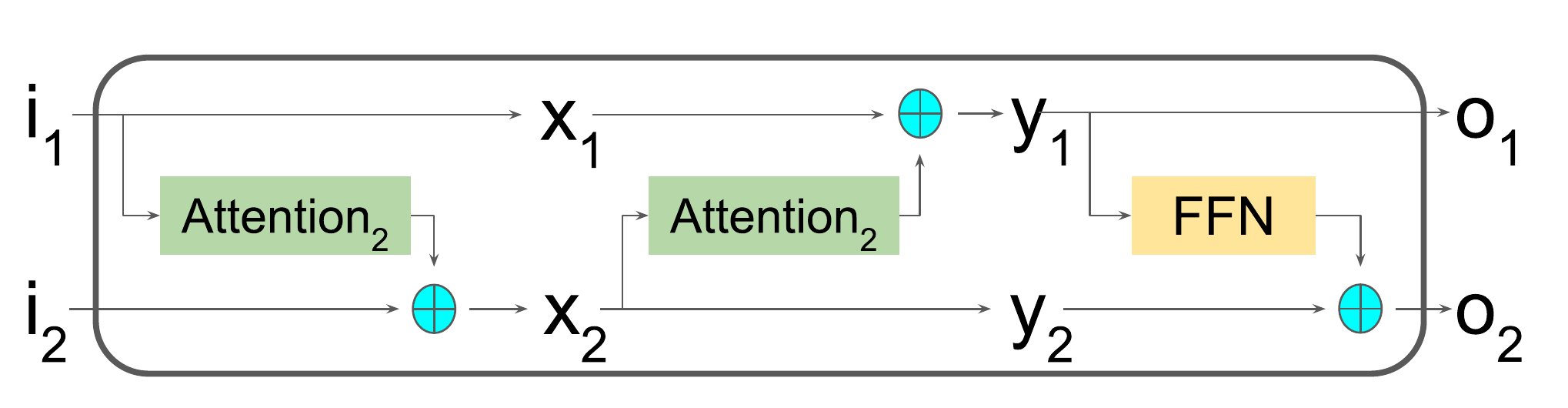}
    \vspace{-0.1cm}
    \caption{Reversible decoder block in Terraformer.}
    \label{fig:terrarev}
\end{figure}

The original Reformer decoder block contained feedforward and attention layers in a 1-1 ratio.
In the Terraformer architecture, as described above, there are two attention layers
in the decoder block, so there are three swaps in the reversible
layers in the decoder block (see Figure~\ref{fig:terrarev}).
In our experiments, this significantly improved performance. 

Another issue with reversibility is that it is only formally correct for continuous functions.
We find that this is not just a formal issue, but an important problem in practice.
To make reversible layers train well with sparsity, we need to store the discrete
decisions---i.e., the integers saying which rows to select---and use them for reversing.
Recalculating these decisions on the backwards pass leads to worse results.

\subsection{Recurrence for Generalization}
\label{sec:terraformer-recurrence}
In addition to incorporating sparse attention and reversibility, we also add recurrence to
the feedforward block of Terraformer. Recurrent layers allow information to propagate in time,
even in a single decoder block. It is challenging though to use them without decreasing model speed, esp. in training. For that reason, we use simple recurrent units \cite{sru} which
parallelize well during training.

SRUs contain dense layers, so their use could negate the benefits
of sparsity elsewhere. We tried a few methods to alleviate that, but it turns out that simply
reducing the dimensionality of the SRUs works. So we first project from $\dm$ to a small
dimension ($32$ in our experiments), then apply the SRU, and then project back to $\dm$ and add
the result to the feedforward block. This low-rank recurrence is in our experiments sufficient
to transfer enough information through time for the network to generalize.

Since the effects of SRUs on C4 are minimal (as the training and evaluation data are very similar),
we use synthetic tasks to investigate out-of-distribution generalization. We train the models on
long addition and on the task of copying a decimal digit. We train on inputs with at most 128 digits
and evaluate on inputs lengths from 256 to 300, so over 2x longer.
As can be seen in the table below, the baseline Transformer does not generalize well, while Terraformer
manages to get a large portion correctly, even if it is not perfect like the Neural GPU \cite{neuralgpu}.

\begin{table}[h]
\begin{center}
\begin{tabular}{r | c c | c c}
\hline
Model & copy & copy (seq) & add  & add (seq) \\
\hline
Transformer   & 79.8\%  & 0\%     & 36.4\% & 0\% \\
Terraformer   & 99.9\%  & 93.9\%  & 86.9\% & 32.4\% \\
\hline
\end{tabular}
\end{center}
\caption{Comparison of out-of-distribution generalization for Terraformer and Transformer on two toy tasks,
  long addition and copying on decimal numbers. Under (seq) we report the number of fully
  correct sequences generated as answers.}
\label{table:r2t2dropout}
\end{table}

\subsection{Experiments}

We designed Terraformer so that the benefits from sparsity would not be lost on long sequences, nor on downstream finetuning tasks. To test this, we chose the task of summarizing scientific papers
using the dataset of scientific papers from arXiv\footnote{We provide full details of
our datasets, hyperparameters, and everything needed to reproduce the results in the appendix. The code is open-sourced as part of Trax 1.4.0 at \url{https://github.com/google/trax}.}\cite{Cohan_2018}.
In this task, the input is a whole paper---a long sequence---and the model is asked to output
its abstract. Several recent papers studied this dataset and tasks and
it has been shown \cite{zhang2020pegasus, zaheer2020big} that pretraining on C4
yields significant improvements on this task.
We also pretrain Terraformer on C4 (like in all experiments in this paper) and fine-tuned it on
the arXiv summarization task.
We find that Terraformer is competitive with the above baselines, even though we mask single words
(we do not use the Pegasus sentence loss) and decode the answers in a greedy way (no beam search). Note that ROUGE scores are computed using open-source scorer\footnote{\url{https://pypi.org/project/rouge-score/}} with the metrics described in its documentation\footnote{\url{https://github.com/google-research/google-research/tree/master/rouge}}. We also observe certain confusion between ROUGE-L metrics reported. As noted in the open-source scorer, there are two versions of ROUGEL-Sentence-Level (R-LSent) and ROUGEL-Summary-Level (R-LSum). For clarity, we report both of these metrics. Furthermore we only report the F1 measure of any ROUGE metric.
We include a few examples of the generated abstracts in the appendix.

\begin{table}
\small
\begin{center}
\begin{tabular}{r  c  c  c c}
 \hline
 Model & R-1 & R-2 & R-LSum & R-LSent \\
\hline
Terraformer        & 45.40 & 17.86  & 41.21 & 26.33 \\
\hline
DANCER RUM         & 42.70 & 16.54 & 38.44 & \textemdash \\
BIGBIRD-RoBERTa    & 41.22 & 16.43 & 36.96 & \textemdash \\
\hline
Pegasus Large (C4) & 44.21 & 16.95 & 38.83 & 25.67 \\
DANCER PEGASUS     & 45.01 & 17.6  & 40.56 & \textemdash \\
BIGBIRD-Pegasus    & 46.63 & 19.02 & 41.77 & \textemdash \\
\hline
\end{tabular}
\end{center}
\caption{Terraformer is competitive with strong baselines \cite{zhang2020pegasus, zaheer2020big, dancerPegasus} on the ArXiv summarization task, without using the Pegasus loss and without beam search.
On R-1, R-2 and R-LSum, Terraformer outperforms all previous models except for BigBird-Pegasus.}
\vspace{-0.1in}
\label{table:arxiv}
\end{table}

We pretrained Terraformer in the same way as all other baselines reported in this paper with the same number of parameters (800M), the same dimensions as mentioned before, and loss sparsity 4 to get the fastest model. Compared to the sparse Transformer model from the previous section that achieves a decoding speed of 0.061s, Terraformer achieves a decoding speed of 0.086s with a similar performance in terms of perplexity (see appendix for details).
We also observe that the Terraformer model achieves accuracy similar to the Transformer model in Table~\ref{table:glue_sparse_ff_qkv} for selected downstream tasks on GLUE dataset.

Table~\ref{table:reformer2big_sparse} shows the speedup in decoding with sparse layers
when we scale up Terraformer to 17B parameters. Note that sparsifying all the layers gives
us 37x speedup in decoding.
% \begin{table}
\begin{wraptable}[12]{L}{0.5\textwidth}
\vspace{-0.5cm}
\begin{center}
\begin{tabular}{r  c  c}
\hline
 Terraformer & Dec. time & Speedup \\
 %    & (17B) & (17B) \\ [0.5ex] 
\hline
dense  & 3.651s & 1x \\ 
 %\hline
Sparse FF & 1.595s & 2.29x\\
 %\hline
SparseFF+QKV & 0.183s & 19.98x \\
 %\hline
SparseFF+QKV+loss & 0.097s & \textbf{37.64x}\\
 \hline
\end{tabular}
\end{center}
\caption{Decoding speed of a single token for Terraformer with 17B parameters is 37x faster than a dense baseline model, requiring less than 100ms/token for inference. Here $\textrm{attention-sparsity}=64$, $\textrm{ff-sparsity}=256$, and $\textrm{loss-sparsity}=4$.}
\label{table:reformer2big_sparse}
\end{wraptable}
% \end{table}

\section{Conclusion}

When starting to investigate sparse variants of Transformers, we assumed that there
would be a price to pay for sparsity---that a sparse model would always
underperform a dense one with the same number of parameters. To our surprise, this is
not the case: sparse is enough! 

In our experiments with large models on the C4 dataset, the sparse models match the performance
of their dense counterparts while being many times faster at inference.
And, when scaling the models up, the benefits of sparsity become even larger.
This promises to put Transformers back on a sustainable track and make large models more useful.

The current results have a number of limitations. For one, the practical speedups we see are only for inference, not at training time. Moreover, we consider unbatched inference on CPUs, while often inference is ran in batched mode on GPUs. We believe with more work sparsity can bring improvements in these settings too,  as our fundamental result shows that the sparse models reach the same perplexity as their dense counterparts with the same number of parameters.

So while we demonstrate that \ourname Transformers are possible, we consider this paper as a first step
on the way to sustainable large models. There are numerous techniques for making models faster
that could greatly benefit Terraformer and other \ourname Transformers. For example, we did not
study quantization and we believe that it can make \ourname Transformers even faster.
We also focused on inference speed and did not get improvements in training speed.
The main reason is our use of Gumbel-Softmax when training the feedforward block
(see Section~\ref{sec:sparse-ff}). \citet{switchtransformer} already provide a
promising alternative, and we look forward to exploring it in future work.

Further, we hope that the community will take inspiration from \ourname Transformers and tune them for their needs.
We ran experiments using layer sizes and hyperparameters borrowed from dense Transformers and they
are most probably not optimal for \ourname Transformer. With proper tuning and further improvements we believe
one could train a \ourname Transformer to match GPT-3 in accuracy but also
run inference in reasonable time on a laptop.
We put it as a fascinating challenge to the community, since such \ourname Transformers will not only be more
sustainable but will also make large models accessible to everyone.

\bibliography{references}
\bibliographystyle{plainnat}

\newpage
\section{Appendix}
\subsection{Sparse QKV}
Sparse QKV uses a multiplicative layer to represent any permutation before composing
this with a convolutional layer. We present the proof that this 
multiplicative layer can represent any permutation below.
\newtheorem*{theorem-nonnumbered}{Theorem}
\begin{theorem-nonnumbered}
With Multiplicative layer defined as
\[
    \textrm{y}_{s, m} = \sum_{i} \textrm{x}_i D_{i,s} E_{i,m}
\]

For any bijective function $f: \{1 \cdots \dm \}\Rightarrow \{1 \cdots S\} \times \{1 \cdots M\}$ there exists a pair of weights of multiplicative layer D, E such that $x_i = y_{s,m}$ for $\{s,m\} = f(i)$.

\end{theorem-nonnumbered}

\begin{proof}
Let's take a function $f$, and define functions $s, m: s(i), m(i) = f(i)$. We construct weights $D_{i,s'} = (1 \text{ if } s'=s(i) \text{ otherwise } 0)$ and $E_{i,m'} = (1 \text{ if } m'=m(i) \text{ otherwise } 0)$. With those constraints we can derive, from the definition of multiplicative layer:
$$ % change to eqnarray*
    \textrm{y}_{s', m'} = \sum_{i} (\textrm{x}_i \text{ if } D_{i,s'}=1 \land E_{i,m'}=1 \text{ otherwise } 0)
$$$$
    \textrm{y}_{s', m'} = \sum_{i} (\textrm{x}_i \text{ if } s'=s(i) \land m'=m(i) \text{ otherwise } 0) \\
$$$$
    \textrm{y}_{s', m'} = \sum_{i} (\textrm{x}_i \text{ if } f(i)=s',m' \text{ otherwise } 0)
$$
Because function $f$ is injective we can use its inversion.
$$
    \textrm{y}_{s', m'} = \sum_{i} (\textrm{x}_i \text{ if } i=f^{-1}(s',m') \text{ otherwise } 0)
$$$$
    \textrm{y}_{s', m'} = \textrm{x}_{f^{-1}(s',m')} \\
$$$$
    \textrm{y}_{f(i)} = \textrm{x}_i
$$
\end{proof} % QED.

\begin{figure}[thb]
    \centering
    \includegraphics[width=8.5cm,keepaspectratio]{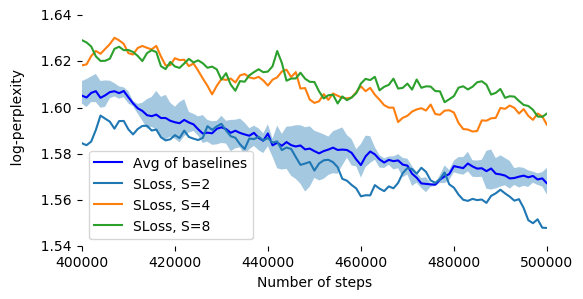}
    \vspace{-0.1in}
    \caption{Log-perplexity of baselines and \ourname Transformers with just Sparse Loss, and varying number of modules.}
    \label{fig:sparse_loss}
\end{figure}

\subsection{Sparse Loss}
To make the loss layer sparse, we investigate the impact of replacing the dense layer with
the multiplicative layer designed for Sparse QKV layer. 
Table~\ref{table:sparse_loss} and Figure \ref{fig:sparse_loss} shows that increasing the sparsity
of the loss layer degrades the perplexity slightly while speeding up the decoding time.

\begin{table}[h]
\begin{center}
\begin{tabular}{r  c  c c} 
\hline
  Sparse loss & Dec. time \\
 %Technique & 500M & 500M  &  \\
 \hline
 baseline  & 0.160 s  \\
% \hline
  S=2  & 0.158 s \\
 %\hline
  S=4  & 0.149 s  \\
 %\hline
  S=8  & 0.148 s  \\
  \hline
\end{tabular}
\end{center}
\caption{Decoding times by varying the number of modules $S$ in sparse loss layer. }
\label{table:sparse_loss}
\end{table}

\subsection{Sparsity Results Data}
The results presented in Figure \ref{fig:intro_sparse_overall} are also accessible via a public Tensorboard link here
\url{https://tensorboard.dev/experiment/on35sXCoTRSoI48ZomOnsw}

\subsection{Architecture for Terraformer}
\begin{figure}
    \centering
    \includegraphics[width=0.6\columnwidth]{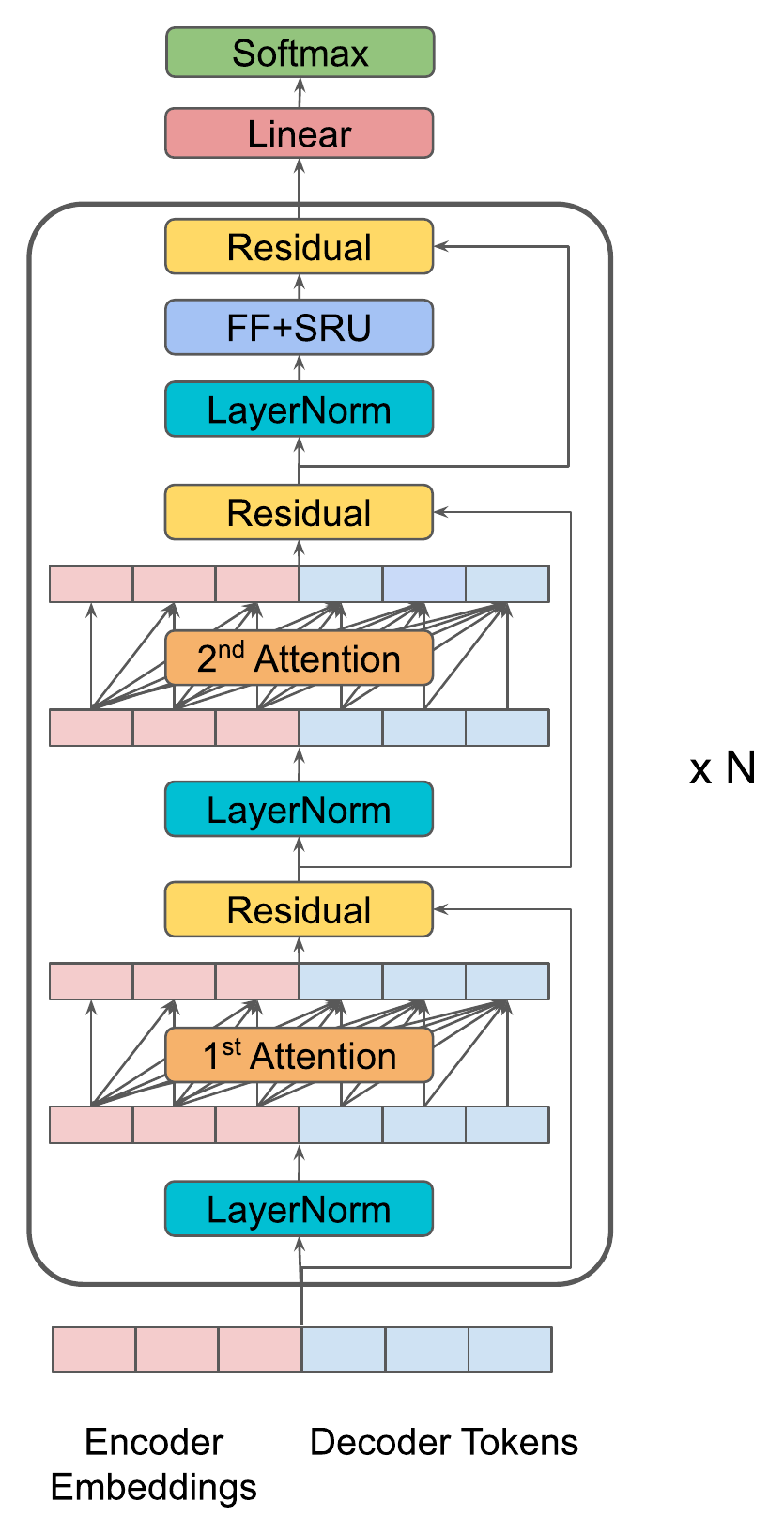}
    \caption{Terraformer Architecture uses two attention mechanisms before the feedforward block in each decoder block.}
    \vspace{-0.1in}
    \label{fig:TerraArch}
\end{figure}

Figure~\ref{fig:TerraArch} shows the whole architecture of Terraformer model discussed in Section~\ref{sec:terraformer-arch}.

\subsection{Pretrained Terraformer on C4 dataset}
We pretrained Terraformer in the same way as all other baselines reported in this paper (see above),
with one difference: we used 4x the batch size. (Thanks to reversibility, Terraformer
can be trained with larger batches.) Table~\ref{table:reformer2medium_sparse} shows the perplexity
and decoding speed of the Terraformer model in comparison to the baseline Transformer model and
the sparse Transformer model from the previous section. All models have the same number
of parameters (800M) and the same dimensions as mentioned before.  We used loss sparsity
4 for Terraformer to get the fastest model, so in Table~\ref{table:reformer2medium_sparse}
we compare it to a sparse Transformer with the same sparse loss.

\begin{table}[h]
\begin{center}
\begin{tabular}{r c c c c c} 
\hline
  & steps & batch size & Log perpl. & Dec. time \\%& Dec. time per block \\
 % & & & per block\\
 \hline
 baseline Transf. & 500k & 4 &  1.57 & 0.160s \\%& 5.9ms \\ 
 sparse Transf. & 500k & 4   &  1.61 & 0.061s \\%& 1.9ms\\
 \hline

 Terraf. & 125k & 16         &  1.66  & 0.086s \\%& 2.9ms \\
 Terraf. & 150k & 16         &  1.63  & 0.086s \\%& 2.9ms \\
 Terraf. & 175k & 16         &  1.59  & 0.086s \\%& 2.9ms \\
  \hline
\end{tabular}
\end{center}
\caption{Terraformer (800M) trained with 4x larger batch size achieves log-perplexity similar to baseline dense Transformer and Scaling Transformers with sparse FF+QKV and sparse loss. Terraformer
trained with larger batch size does not match the perplexity of the baseline at $\frac{1}{4}$th number
of steps, but catches up at around $\frac{1}{3}$rd---we believe this may be due to the fact that we used
training hyperparameters optimized for the baselines. Decoding of a single token is 1.92x faster than baseline.}
\label{table:reformer2medium_sparse}
\end{table}

\subsection{Finetuning Terraformer on summarization task}
We present a few examples of the abstracts generated by the Terraformer model for scientific papers in the ArXiv dataset \cite{Cohan_2018}. Table~\ref{table:AbstractExamples} compares these abstracts to the corresponding examples from Tables~I.{25-27} in section~I of \cite{zhang2020pegasus}. 

The abstracts are decoded using greedy algorithm with temperature $T=0.5$.

\onecolumn
\begin{longtable}[]{p{0.1\linewidth} | p{0.85\linewidth}}
\small

& ArXiv \\
\hline
Document (ID \#34)   & consider a set of objects which should be ranked on the basis of information about their bilateral relationships . similar problems arise , among others
, in social choice theory xcite , sports xcite , psychology xcite , internet search xcite , and bibliometrics xcite . we discuss a universal version of
the problem involving arbitrary preference intensities as well as incomplete and multiple comparisons . the main contribution of this paper is the
presentation of an impossibility theorem : consistency requiring that if an object is ranked at least as high as another in two independent problems
, then it is ranked as high as the other in the unified problem , too and self - consistency a less known but intuitive property , introduced in xcite
, which prohibits to assign a lower rank for an object with a better or equivalent performance than another can not be met simultaneously by any
ranking method on the set of all problems . domain restrictions and weakening of the properties are also investigated in order to get some positive
results . since self - consistency is an axiom difficult to debate , we can not demand consistency from a ranking method . at the first glance , it is a
somewhat surprising and controversial result making the aggregation of ( incomplete ) preferences impossible . one may argue that the whole concept
of paired comparisons - based ranking is flawed . however , loss of consistency may be regarded as a necessary sacrifice for dimension reduction
since the outcome should be an inherently transitive ranking even from intransitive data . this interpretation is reinforced by the connection between
consistency and independence of irrelevant matches ( which requires the relative ranking of two objects to be independent of the outcome of ... ... ... \\
\hline
Golden & assume a set of objects is given with information about their bilateral relationships , allowing for incomplete and multiple comparisons as well as
different preference intensities . an axiomatic approach is applied for the problem of ranking the objects . consistency requires the preservation
of relative ranking if two sets of such preferences are aggregated . self - consistency assigns the same rank for objects with the same performance
, furthermore , an object should be ranked strictly higher if it shows an obviously better performance than another . it is revealed that these two
properties can not be satisfied simultaneously . the impossibility holds under various restrictions on the domain . however , a positive result emerges
if only the aggregation of ranking problems with the same comparison structure is allowed , for example , we have two round - robin tournaments . \\
\hline
Terraformer & we discuss a universal version of the problem involving arbitrary preference intensities as well as incomplete and multiple comparisons . the main contribution of this paper is the presentation of an impossibility theorem : consistency requiring that if an object is ranked at least as high as another in two independent problems , then it is ranked as high as the other in two independent problems , too and self - consistency a less known but intuitive property , introduced in @xcite . first , independence of irrelevant matches ( posing a kind of independence concerning the relative ranking of two objects ) is implied by three independent axioms , neutrality ( independence of relabelling the objects ) , symmetry ( implying a flat ranking if all comparisons are drawn ) , and consistency ( perhaps the most natural property of additivity ) . it turns out that consistency and self - consistency are still enough to get this result . the size of the problem does not solve the issue as it emerges with at least four objects . the size of the problem does not solve the issue as it emerges with at least four objects . the size of the problem is not a trivial extension of the aggregation of ( incomplete ) preferences .\\
\hline
ROUGE-1 & 41.87 \\
\hline
\hline
Document (ID \#289)   & machine learning methods are used widely within high energy physics ( hep ) . one promising approach , used extensively outside of hep for
applications such as handwriting recognition , is that of support vector machines ( svms ) , a supervised learning model used with associated learning
algorithms for multivariate analysis ( mva ) . developed originally in the 1960s , with the current standard version proposed in 1995 xcite , svms aim
to classify data points using a maximal margin hyperplane mapped from a linear classification problem to a possibly infinite dimensional hyperspace
. however this means svms , like other mva classifiers , have a number of free parameters which need to be tuned on a case by case basis . this
motivates a number methods for ensuring the classifier is sufficiently generalised such that when used on an unseen dataset the performance can be
accurately predicted . in this paper a brief overview of svms is given in section [ sec : svm ] , with an example using svms shown in section [ sec :
checker ] . generalisation is discussed in section [ sec : gen ] with an illustrative example of how this can improve performance given in section [ sec
: check2 ] . consider the problem of linear classification with the svm where the training set , xmath0 , is linearly separable . we define a separating
hyperplane given by xmath1 , where xmath2 , the weight vector , is perpendicular to the hyperplane , and xmath3 , the bias , determines the distance
of the hyperplane from the origin ( fig . ) . a decision function defined by xmath4 is used to positively and negatively classify xmath5 , the points in
the training set . without further constraint the ... ... ...\\
\hline
Golden & we review the concept of support vector machines ( svms ) and discuss examples of their use . one of the benefits of svm algorithms , compared
with neural networks and decision trees is that they can be less susceptible to over fitting than those other algorithms are to over training . this issue
is related to the generalisation of a multivariate algorithm ( mva ) ; a problem that has often been overlooked in particle physics . we discuss cross
validation and how this can be used to improve the generalisation of a mva in the context of high energy physics analyses . the examples presented
use the toolkit for multivariate analysis ( tmva ) based on root and describe our improvements to the svm functionality and new tools introduced for
cross validation within this framework .\\
\hline
Terraformer & support vector machines ( svms ) are used widely within high energy physics ( hep ) for applications such as handwriting recognition . the svm algorithm is a supervised learning model used with associated learning algorithms for multivariate analysis ( mva ) . the hard margin svm is an important svm algorithm , and the hard margin svm is an important svm algorithm . in this paper a brief overview of svms is presented , with an example showing similar performance to that of a bdt . however it is not clear without further checks as to whether the mvas are sufficiently generalised . hence a multistage cross - validation procedure has been outlined , which for the same example shows better performance as well as better agreement between the training and testing samples in the output distributions .\\
\hline
ROUGE-1 & 43.54\\
\hline
\hline
Document (ID \#294)   & the pure quantum states are traditionally associated with the wave function xcite or a vector in the hilbert space xcite . the mixed quantum states
are described by the density matrix xcite or the density operator xcite . there exist several representations of quantum states in terms of the quasidistribution functions like the wigner function xcite and the husimi kano function xcite . the diagonal representation of quantum states was suggested
in xcite ( see also xcite ) . it was studied and applied in xcite . in this representation , a quantum state is represented in terms of weighted sum
of coherent - state xmath0 projectors . the properties of all the quantum - state representations considered are associated with the properties of the
density operator which is hermitian , trace - class nonnegative operator . this means , in particular , that all the eigenvalues of the density operators
must be nonnegative . in the quantum domain , the multipartite systems have a specific property connected with strong correlations of the quantum
subsystems . this property provides the entanglement phenomenon xcite . in the diagonal representation of the density states , the weight function
xmath1 is an analog of the probability - distribution function in the phase space . for some class of states , this function is identical to the probability
- distribution function like in classical statistical mechanics . in xcite , the tomographic - probability representation of quantum states , where the
quantum state is associated with the so - called symplectic tomogram , was introduced . the tomogram is a fair probability distribution containing the
same information on quantum state that the density operator does ( or such its characteristics as the wigner or husimi kano functions ) . the aim of
this work ... ... ...\\
\hline
Golden & the quasidistributions corresponding to the diagonal representation of quantum states are discussed within the framework of operator - symbol
construction . the tomographic - probability distribution describing the quantum state in the probability representation of quantum mechanics is
reviewed . the connection of the diagonal and probability representations is discussed . the superposition rule is considered in terms of the density
- operator symbols . the separability and entanglement properties of multipartite quantum systems are formulated as the properties of the density -
operator symbols of the system states .\\
\hline
Terraformer & the diagonal and probability representations of quantum states are reviewed . the superposition rule is formulated in both the diagonal and probability representations . the problem of separability and entanglement is formulated in both the diagonal and probability representations . the problem of separability and entanglement is formulated in both the diagonal and probability representations . the problem of separability and entanglement is formulated in both the diagonal and probability representations . the separability and entanglement properties of composed system in the tomographic and diagonal representations are analyzed .\\
\hline
ROUGE-2 & 27.16\\
\hline
\hline
Document (ID \#32)   & deep neural networks ( dnns ) have been receiving ubiquitous success in wide applications , ranging from computer vision xcite , to speech recognition
xcite , natural language processing xcite , and domain adaptation xcite . as the sizes of data mount up , people usually have to increase the number
of parameters in dnns so as to absorb the vast volume of supervision . high performance computing techniques are investigated to speed up dnn
training , concerning optimization algorithms , parallel synchronisations on clusters w / o gpus , and stochastic binarization / ternarization , etc xcite
. on the other hand the memory and energy consumption is usually , if not always , constrained in industrial applications xcite . for instance , for
commercial search engines ( e.g. , google and baidu ) and recommendation systems ( e.g. , netflix and youtube ) , the ratio between the increased
model size and the improved performance should be considered given limited online resources . compressing the model size becomes more important
for applications on mobile and embedded devices xcite . having dnns running on mobile apps owns many great features such as better privacy , less
network bandwidth and real time processing . however , the energy consumption of battery - constrained mobile devices is usually dominated by
memory access , which would be greatly saved if a dnn model can fit in on - chip storage rather than dram storage ( c.f . xcite for details ) . a recent
trend of studies are thus motivated to focus on compressing the size of dnns while mostly keeping their predictive performance xcite . with different
intuitions , there are mainly two types of dnn compression methods , which could be used in conjunction for better parameter savings . the ... ... ...\\
\hline
Golden & as the complexity of deep neural networks ( dnns ) trend to grow to absorb the increasing sizes of data , memory and energy consumption has
been receiving more and more attentions for industrial applications , especially on mobile devices . this paper presents a novel structure based on
functional hashing to compress dnns , namely funhashnn . for each entry in a deep net , funhashnn uses multiple low - cost hash functions to fetch
values in the compression space , and then employs a small reconstruction network to recover that entry . the reconstruction network is plugged into
the whole network and trained jointly . funhashnn includes the recently proposed hashednets xcite as a degenerated case , and benefits from larger
value capacity and less reconstruction loss . we further discuss extensions with dual space hashing and multi - hops . on several benchmark datasets ,
funhashnn demonstrates high compression ratios with little loss on prediction accuracy .\\
\hline
Terraformer & this paper presents a novel approach to relieve the hash collision effect through a sparse feature space . specifically , we use multiple hash functions to map per virtual entry into multiple values in compression space . then an additional network plays in a \_ mapping function \_ role from these hashednets to the virtual entry before hashing , which can be also regarded as  \u2047  reconstructing '' the virtual entry from its multiple hashed values . plugged into and jointly trained within the original network , the reconstruction network is of a comparably ignorable size , i.e. , at low memory cost . this functional hashing structure includes hashednets as a degenerated special case , and facilitates less value collisions and better value reconstruction . experiments on several datasets demonstrate promisingly larger reduction of model sizes and/or less loss on prediction accuracy , compared with hashednets .\\
\hline
ROUGE-2 & 16.11\\
\hline
\hline
Document (ID \#248)   &  stripped supernovae ( sne ) and long - duration gamma - ray bursts ( long grbs ) are nature s most powerful explosions from massive stars . they
energize and enrich the interstellar medium , and , like beacons , they are visible over large cosmological distances . however , the mass and
metallicity range of their progenitors is not known , nor the detailed physics of the explosion ( see reviews by xcite and xcite ) . stripped - envelope
sne ( i.e , sne of types iib , ib , and ic , e.g. , xcite ) are core - collapse events whose massive progenitors have been stripped of progressively larger
amounts of their outermost h and he envelopes ( fig . [ fig1 ] ) . in particular , broad - lined sne ic ( sne ic - bl ) are sne ic whose line widths approach
20,000xmath030,000 xmath1 around maximum light ( see below ) and whose optical spectra show no trace of h and he . for the last 15 years , the
exciting connection between long grbs and sne ic - bl , the only type of sne observed accompanying long grbs ( for reviews , see xcite ) , and the
existence of many more sne ic - bl without grbs raises the question of what distinguishes sn - grb progenitors from those of ordinary sne ic - bl
without grbs . viewing angle effects are probably not the reason why those sne ic - bl did not show an accompanied grbs xcite and based the same
radio upper - limits , only xmath2 1\% of sne ib / c appear to be accompanied by grbs xcite . one promising line of attack is to investigate what sets
apart sne ib ... ... ...\\
\hline
Golden & while the connection between long gamma - ray bursts ( grbs ) and type ib / c supernovae ( sne ib / c ) from stripped stars has been well - established
, one key outstanding question is what conditions and factors lead to each kind of explosion in massive stripped stars . one promising line of attack
is to investigate what sets apart sne ib / c * with * grbs from those * without * grbs . here , i briefly present two observational studies that probe
the sn properties and the environmental metallicities of sne ib / c ( specifically broad - lined sne ic ) with and without grbs . i present an analysis
of expansion velocities based on published spectra and on the homogeneous spectroscopic cfa data set of over 70 sne of types iib , ib , ic and ic -
bl , which triples the world supply of well - observed stripped sne . moreover , i demonstrate that a meta - analysis of the three published sn ib /
c metallicity data sets , when including only values at the sn positions to probe natal oxygen abundances , indicates at very high significance that
indeed sne ic erupt\\
\hline
Terraformer & the discovery of stripped supernovae ( sne ) and long - duration gamma - ray bursts ( grbs ) , and the discovery of long - duration gamma - ray bursts ( grbs ) , has raised the question of what distinguishes sne ic - bl from those of ordinary sne ic - bl without grbs . here i present a meta - analysis of published metallicities at the explosion site of stripped supernovae with and without grbs , and a meta - analysis of published metallicities at the explosion site of sne ib / c with and without grbs . the results of our analysis are compared with those of sne ic - bl without grbs , and with those of sne ic - bl without grbs . the results of our study are compared with those of sne ic - bl without grbs , and with sne ic - bl with grbs .\\
\hline
ROUGE-LSent & 28.83\\
\hline
\hline
Document (ID \#270)   & the alice experiment is dedicated to the study of the properties of qcd matter created in nucleus - nucleus collisions at lhc energies xcite . the inner
tracking system in the alice apparatus is made of position sensitive detectors which have to handle several thousands tracks per unit of rapidity . the
two innermost layers at 3.9 xmath10 and 7.6 xmath10 radii , respectively , constitute the silicon pixel detector ( spd ) . the spatial precision and hit
efficiency of the spd are key parameters since they determine the alice capability of detecting particles with open heavy - flavour xcite . + the basic
detector unit of the alice spd is the ladder , a two - dimensional silicon matrix of pxmath11n reverse biased diodes of dimensions 50 x 425 xmath12 ,
flip - chip bonded to five read - out chips . each diode is connected to a cell of the front - end read - out asic via a pb - sn solder bump of 25 xmath13
diameter . the detector contains nearly 10xmath14 active cells in total . the read - out is binary . to reduce the material budget , the sensor thickness
is limited to 200 xmath13 and the read - out chip wafers are thinned down to 150 xmath13 . further details can be found in xcite . + early prototypes
of the alice spd elements , in the form of single - chip assemblies , were tested in high energy proton / pion beams at the cern sps in 2002 and 2003 .
these assemblies were made with sensors of 200 xmath13 and 300 xmath13 thicknesses , while the read - out chips ( unthinned ) were 725 xmath13
thick . those beam tests were primarily aimed at evaluating the performance of ... ... ...\\
\hline
Golden & the two innermost layers of the alice inner tracking system are instrumented with silicon pixel detectors . single chip assembly prototypes of the alice pixels have been tested in high energy particle beams at the cern sps . detection efficiency and spatial precision have been studied as a function of the threshold and the track incidence angle . the experimental method , data analysis and main results are presented . d. elia@xmath0 , g. anelli@xmath1 , f. antinori@xmath2 , a. badal@xmath3 , g.e . bruno@xmath4 , m. burns@xmath1 , i.a . cali@xmath5 , m. campbell@xmath1 , m. caselle@xmath4 , s. ceresa@xmath1 , p. chochula@xmath1 , m. cinausero@xmath6 , j. conrad@xmath1 , r. dima@xmath2 , d. fabris@xmath2 , r.a . fini@xmath4 , e. fioretto@xmath6 , s. kapusta@xmath1 , a. kluge@xmath1 , m. krivda@xmath7 , v. lenti@xmath4 , f. librizzi@xmath3 , m. lunardon@xmath2 , v. manzari@xmath4 , m. morel@xmath1 , s. moretto@xmath2 , p. nilsson@xmath1 , f. osmic@xmath1 , g.s . pappalardo@xmath3 , v. paticchio@xmath4 , a. pepato@xmath2 , g. prete@xmath6 , a. pulvirenti@xmath3 , p. riedler@xmath1 , f. riggi@xmath3 , l. sndor@xmath7 , r. santoro@xmath4 , f. scarlassara@xmath2 , g. segato@xmath2 , f. soramel@xmath8 , g. stefanini@xmath1 , c. torcato de matos@xmath1 , r. turrisi@xmath2 , l. vannucci@xmath6 , g. viesti@xmath2 , t. virgili@xmath9 spatial precision , efficieny , silicon pixel detector , alice , lhc 29.40.gx , 29.40.wk\\
\hline
Terraformer & the performance of prototype assemblies for the alice silicon pixel detector have been extensively studied by using beam test data collected in the past years at the cern sps . the cluster pattern distribution , the intrinsic spatial precision and the detection efficiency have been investigated as a function of both the detector threshold and the incident angle of the tracks . the results show a very high detection efficiency ( above 99\% ) in a wide threshold range and a spatial precision of about 10 @xmath0 in the short pixel side coordinate for normal track incidence and 210 dac threshold . the detector performance with angled tracks has also been investigated .\\
\hline
ROUGE-LSent & 23.10\\
\hline

%\end{center}
\caption{Abstracts generated by the Terraformer model on a set of examples of scientific papers from ArXiv. Note that the corresponding abstracts and scores for the Pegasus model are in  Tables~I.{25-27} in section~I of \cite{zhang2020pegasus}.}
\label{table:AbstractExamples}
\end{longtable}
\twocolumn

\end{document}